\documentclass[letterpaper, 10 pt, conference]{ieeeconf}  

\IEEEoverridecommandlockouts                              

\usepackage{graphics} 
\usepackage{epsfig} 
\usepackage{mathptmx} 
\usepackage{times} 
\usepackage{amsmath} 
\usepackage{amssymb}  
\usepackage{xspace}
\usepackage{paralist}

\usepackage{subfig}
\usepackage{makecell}
\usepackage{url}
\usepackage{colortbl}
\usepackage{multirow}

\newcolumntype{P}[1]{>{\centering\arraybackslash}p{#1}}

\newcommand{\mypartitle}[2][2.]{\vspace*{-#1 ex}~\\{\noindent {\bf #2}}}

\newcommand{\liftedpose}[0]{\ensuremath{\bar{\mathbf x}}\xspace}
\newcommand{\truepose}[0]{\ensuremath{\mathbf x^{*}}\xspace}
\newcommand{\totalpose}[0]{\ensuremath{\hat{\mathbf x}}\xspace}

\newcommand{\NJoints}[0]{\ensuremath{J}\xspace}
\newcommand{\NFeatures}[0]{\ensuremath{2048}\xspace}
\newcommand{\RegStages}[0]{\ensuremath{S}\xspace}

\newcommand{\ResPose}[0]{R-Pose\xspace}
\newcommand{\CoordReg}[0]{C-Reg\xspace}

\def\idiapref{\textsuperscript{\textasteriskcentered}}
\def\epflref{\textsuperscript{\textdagger}}

\makeatletter
\DeclareRobustCommand\onedot{\futurelet\@let@token\@onedot}
\def\@onedot{\ifx\@let@token.\else.\null\fi\xspace}

\def\eg{{e.g}\onedot}

\makeatother

\definecolor{Gray}{gray}{0.9} 

\title{\LARGE \bf
Residual Pose: A Decoupled Approach for Depth-based \\ 3D Human Pose Estimation
}

\author{Angel Mart\'inez-Gonz\'alez\idiapref\epflref, Michael Villamizar\idiapref, Olivier Can\'evet\idiapref\ and Jean-Marc Odobez\idiapref\epflref
\thanks{$^{*}$ Idiap Research Institute, Switzerland. \{angel.martinez, michael.villamizar, olivier.canevet, odobez\}@idiap.ch }
\thanks{\textsuperscript{\textdagger} \'Ecole Polytechnique F\'ed\'erale de Lausanne (EPFL), Switzerland.}
}

\begin{document}

\maketitle
\thispagestyle{empty}
\pagestyle{empty}

\begin{abstract}
%
%
We propose to leverage recent advances in reliable 2D pose estimation with
Convolutional Neural Networks (CNN) to estimate the 3D pose of people
from depth images in multi-person Human-Robot  Interaction (HRI) scenarios.
Our method is based on the observation that using the depth information
to obtain 3D lifted points from 2D body landmark detections
provides a rough estimate of the true 3D human pose,
thus requiring only a refinement step.
In that line our contributions are threefold.
(i) we propose to perform 3D pose estimation from depth images by 
decoupling 2D pose estimation and 3D pose refinement;
%
(ii) we propose a deep-learning approach that regresses the residual pose
between the lifted 3D pose and the true 3D pose;
(iii) we show that  despite its simplicity, our approach achieves very
competitive results both in accuracy and speed on two public datasets
and is therefore appealing for multi-person HRI
compared to recent state-of-the-art methods.

\end{abstract}

\section{Introduction}\label{sec:intro}
%
%
%
3D human pose estimation is an essential part of many applications involving
human behavior analysis, like  3D scene understanding, social robotics,
visual surveillance and gaming.
For instance,  in social HRI, the ability to sense the  3D pose
of humans provides to the robot the means to further recognize their activity
or evaluate their interaction engagement.
However, although  3D pose estimation has been a very important  topic of research,
factors like person self occlusions, pose variations, sensing conditions and
low computational budget
increase the challenge of deploying accurate, reliable and efficient
3D pose estimation systems.

\mypartitle{State-of-the-art.}
Early approaches on 3D human pose estimation 
detect body landmarks in the image that are then coupled with 3D human
pose priors that account for body kinematics and physical 
constraints~\cite{Sigal:IJCV:11,Ramakrishna_ECCV_2012}.
Nowadays, Deep Neural Networks (DNN) have become the mainstream approach,
which lead to the emergence of a large number of methods to address 3D  pose estimation from 
color~\cite{Chen_2017_CVPR,martinez_2017_3dbaseline,Tome_CVPR_2017}
and depth images~\cite{DepthInvariant,pavlakos2017volumetric,Moon_2018_CVPR_V2V-PoseNet,ShottonPAMI,Taylor_Viturvian_CVPR_2018}.
%
%

%
From a methodological perspective, methods can nevertheless be grouped into two main 
threads: fitting and learning methods.
The former ones extend earlier works, but rather use CNNs to localize 2D body parts,
and then fit a 3D body pose model along with constraints  via an 
optimization objective defined in the image domain~\cite{KEEPSMPL_ECCV_2016}
or in the  2D-to-3D joint space~\cite{Li_ICCV_2015,Chen_2017_CVPR,Tome_CVPR_2017}.
Learning based methods take advantage of the recent DNNs
to directly predict and regress the 3D locations of the body parts with
fully connected networks~\cite{Habibie_CVPR_2019,martinez_2017_3dbaseline}.
%
%
Although simple, the 3D coordinate regression has proved to be an effective and
efficient  solution.
%
Moreover, information about pose kinematics can  been incorporated
as an additional limb loss~\cite{Sun2017CompositionalHP}, 
using a structured prediction layer~\cite{Aksan_2019_ICCV}, 
or via a re-projection regularizer~\cite{kanazawaHMR_CVPR_18,Wandt_CVPR_2019}.
However, a drawback of these models is that they predict the 3D coordinates
with respect to a root joint that is assumed to be known in advance, or which in
practice needs to be predicted as well.

\begin{figure}[t]
\centering
\includegraphics[width=0.75\linewidth]{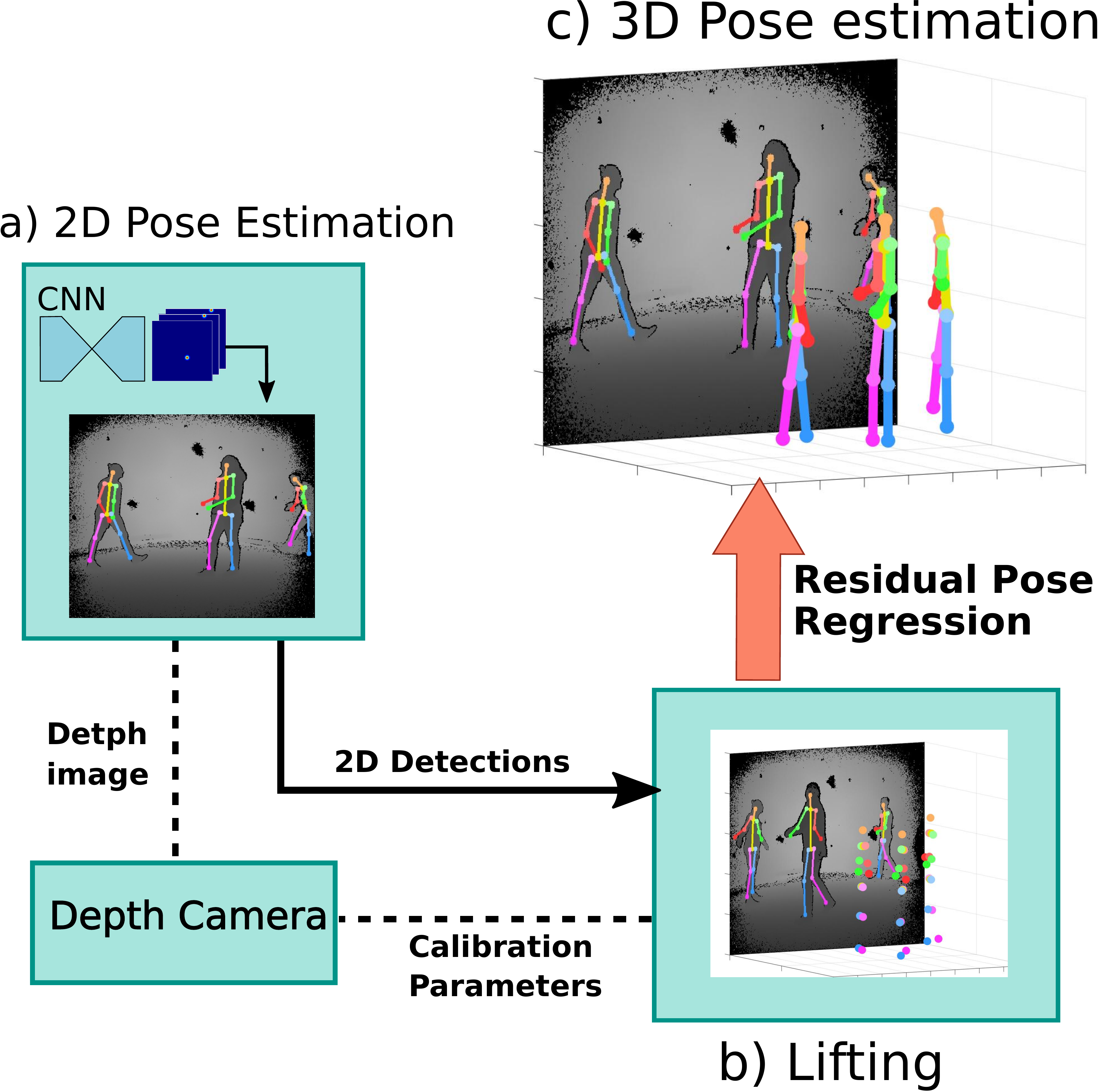}

	\vspace{-0.1cm}
\caption{Proposed decoupled residual pose  approach: 
		a) bottom-up multi-person 2D pose detection;
		b) for each detected person, 2D body joints are lifted to the 3D space.
		c) 3D pose estimation using a residual pose regression network.
}
\label{fig:pipeline}
\end{figure}

The depth data modality has also been largely exploited, since compared to
color images it is texture and color invariant,
and helps to remove the ambiguities in scale and shape
by providing direct access to 3D information.
As with color, some methods tend to rely on fitting approaches,
for instance by  identifying one-to-one relationships between cloud points and a 
3D mesh via Iterative Closest Point (ICP)~\cite{ICP_ICCV_2011}
or using random forest~\cite{Taylor_Viturvian_CVPR_2018}.
Other approaches model the 3D location distributions of 
3D points with respect to their parents in a kinematic tree~\cite{Jung_RTW_CVPR_2015,ShottonPAMI}.
As a typical example, the seminal work of Shotton~\cite{ShottonPAMI}
employed random forests to classify depth pixels into  different human joints
and used weighted voting to estimate their  3D locations.
%
%
Deep learning also improved upon these works~\cite{DepthInvariant,Moon_2018_CVPR_V2V-PoseNet,guo2017towards,DepthMultitask}.
In \cite{DepthInvariant}, multi-view human pose estimation is solved by learning a view invariant feature space
and iteratively refining  the 3D coordinates with a Recurrent Neural Network (RNN).
In~\cite{Moon_2018_CVPR_V2V-PoseNet} depth images are transformed into a voxelized representation
and 3D Gaussian likelihoods are predicted for each body joint per voxel using a costly 3D-CNN.
%
However,  these methods usually work on image crops centered around the person.
As a consequence, to handle the multi-person case, 
a person detector is still needed as a first step,
followed by multiple forward passes of a relatively heavy image processing network to estimate
the 3D pose of each detected person, leading to an  increased computational cost.
%
%

%
\mypartitle{Approach and contributions.}
%
%
%
An overview of our approach for accurate and fast multi-person 3D pose estimation
is presented in Fig.~\ref{fig:pipeline}.
Our main idea is to better exploit the depth information and decouple the task in
two main steps:
2D multi-person pose estimation and 3D pose regression.
The motivations are that the first step can benefit from
recent accurate and efficient architectures to achieve this task,
and that the second one can be done efficiently by directly
regressing the 3D pose coordinates from the 2D ones in
two substeps: a simple but effective scheme which lifts the 2D estimates
to 3D using the depth information and pose priors (to handle partial occlusion);
and a novel efficient residual pose 3D regression methods  that works on this set of points.
%
%
%
%
This makes our approach computationally ligher for multi-person HRI settings
since compared to CNNs applied to image crops for 3D pose prediction, 
the cost of our 3D regression scheme is much smaller, and the cost saving is proportional
to the number of person in the scene.
%
%
In this context, our contributions can be summarized as:
\begin{compactitem}
\item we investigate an innovative method decoupling the 3D pose estimation task 
  into an accurate and efficient CNN-based 2D bottom-up 
  multi-person pose estimation method and 3D pose regression;
  %
  %
\item we propose a simple 2D-to-3D lifting scheme which handles 2D body joint miss detections;
  %
\item we introduce  a novel method for 3D pose regression
  from lifted 2D estimates by relying on a residual-pose
  deep-learning architecture;
  %
  %
\item we demonstrate that despite its simplicity, 
  our approach achieves very competitive results
  on different public datasets and is suitable for multi-party HRI scenarios.
  %
\end{compactitem}
Models and code will be made  publicly
available\footnote{\color{red}\url{https://github.com/idiap/residual_pose}}.
%
The paper is organized as follows: Section~\ref{sec:cnns}
introduces our strategy for 2D pose estimation and lifting.
Section~\ref{sec:3dpose} presents our approach and regressor neural network architecture
for residual pose learning.
%
Experiments are described in Section~\ref{sec:experiments}, 
and Section~\ref{sec:conclusion} presents our conclusions.

\section{Efficient 2D Pose Estimation and Lifting}
\label{sec:cnns}
%
%
%
%
%
%
%
This section describes the CNN architectures used for accurate bottom-up 2D pose
estimation and our proposed  method for 2D-to-3D body joint lifting and
for handling  miss-detections due to (self-)occlusion or failures.

\subsection{CNN-based 2D Pose Estimation}
We follow recent breakthroughs in multi-person 2D pose estimation 
that use a CNN to predict confidence maps $\rho(\cdot)$ for the location of the body 
landmarks in the image and part affinity fields $\phi(\cdot)$ for the location and orientation of the limbs~\cite{CPMPaf}.
We analyze different CNN architectures and the impact of their 2D
estimates on the quality of the 3D pose.

\begin{figure}[t]
\centering
\includegraphics[width=\linewidth]{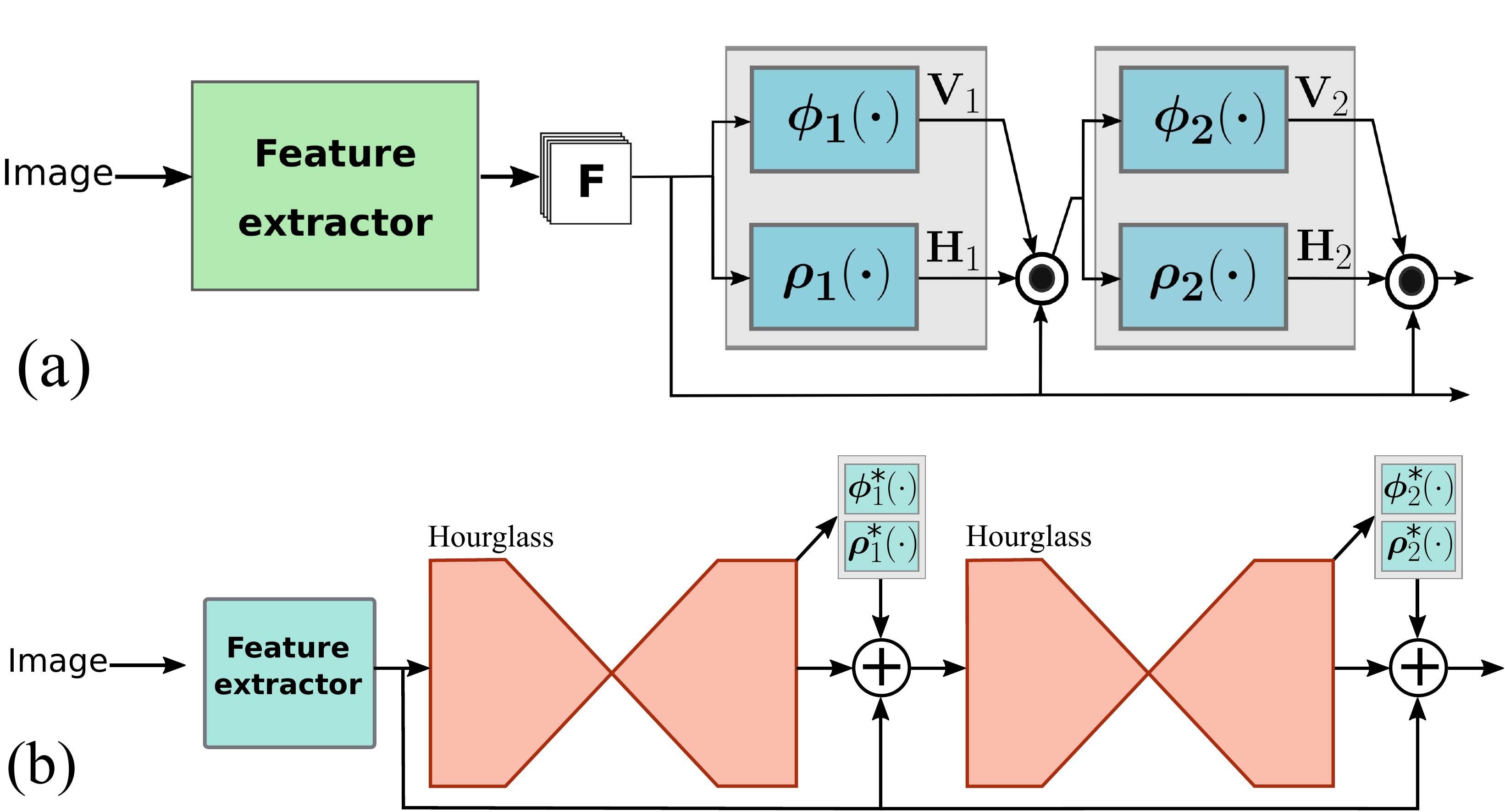}
\vspace{-0.2cm}
\caption{CNN architectures used for  2D pose estimation.
  %
  (a)~Pose Machine architecture implemented by RPM and MPM~\cite{Martinez_TCSVT_2019}.
  (b)~Our extension of the Hourglass network for multi-person 2D pose estimation.
}
\label{fig:2d-CNN}
\vspace{-2mm}
\end{figure}

Three architectures are considered. The two firsts are the efficient pose machines
based on residual  modules (RPM) and the one based on MobileNets (MPM) introduced in~\cite{Martinez_TCSVT_2019}.
These are lightweight CNNs that refine predictions with a series
of prediction stages and are designed for efficient 2D pose estimation 
with real-time performance, see Fig.\ref{fig:2d-CNN}~(a).
Additionally, we consider the Hourglass network architecture~\cite{HourGlass}
which was originally proposed for single person pose estimation.
It comprises a series of UNet-like networks that process image features at 
different semantic levels.
We follow the original design  but adapt the output
to predict part affinity fields to match our multi-person scenario
by branching a duplicate of the confidence maps prediction layers
(Fig.\ref{fig:2d-CNN} (b)).

\subsection{Pose lifting}
\label{sec:lifting}
Given 2D landmark detections, we use their corresponding depth values $Z$
to lift them according to $\bar{x} = Z \cdot K \cdot (x_{img}, y_{img}, 1)^{\top}$,
%
where $K= diag( 1/f_x, 1/f_y,1)$ is the depth camera matrix.
However, different errors can arise.
%
For example, a 2D detection might have missing depth value due to sensing failures.
Additionally, as is common in typical HRI scenarios, self and between-person 
occlusions will naturally result in missing body detections.

\begin{figure}[t]
\includegraphics[width=\linewidth]{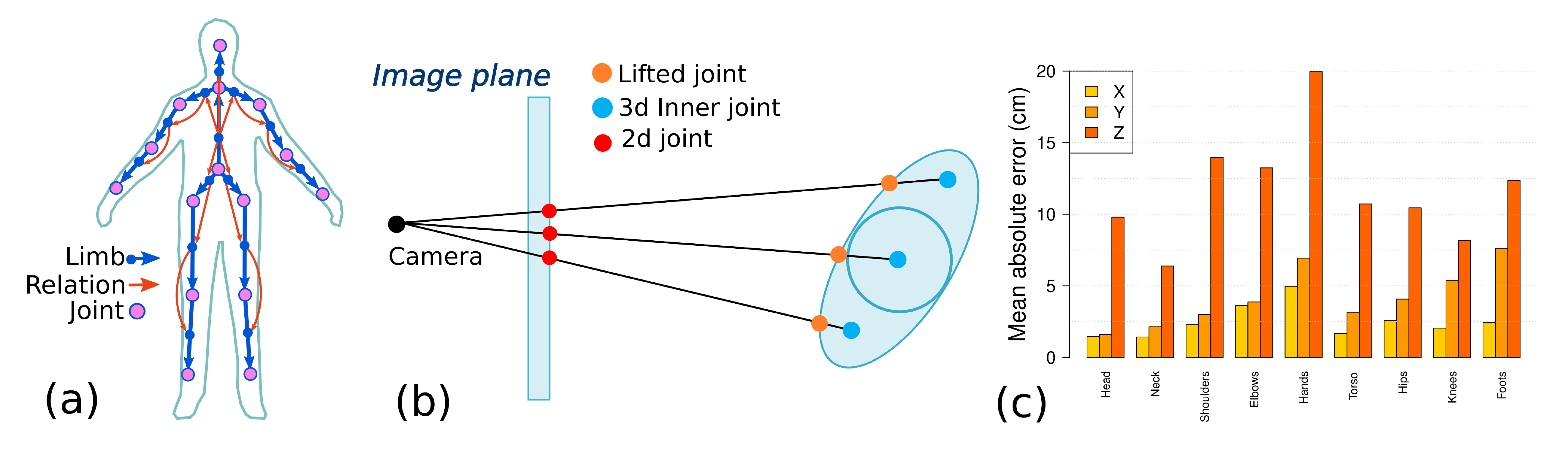}

\vspace{-0.2cm}
\caption{
  (a) Skeleton and limb pairwise relationships;
  (b) illustration of the error introduced by the lifting process of the 2D detected landmarks;
  (c) mean absolute error on each coordinate when using
  the 3D lifted points as the 3D estimation on the ITOP dataset.
}
\vspace{-2mm}
\label{fig:calibrated-skeleton}
\end{figure}

In this paper, in these cases, rather than feeding our regressor with dummy values 
which might bias estimations, we propose a simple  recovery method.
First, in case of missing depth values,
we use the mean depth of the points with valid depth information
in the landmark's vicinity.
Second, in case of missed landmark detections, we rely on a 3D pose prior to
infer their expected coordinates.
However, rather than relying on expensive-to-compute prior~\cite{Sigal:IJCV:11}, we follow a 
simpler 3D limb prior based on  pairwise relationships between limb vectors.
Following a tree of limbs from the skeleton and taking the spine limb as root (see Fig.~\ref{fig:calibrated-skeleton}(a)),
we consider adjacent limbs, encode their 3D direction and length within a joint Gaussian distribution
$p(l_i, l_{\mathbf{pa}(l_i)})$, and learn the model parameters from training data.
Then, to predict the lifted coordinates $\bar{x_i}$ of a missed landmark,
we consider its associated limb $l_i$ in the skeleton whose other landmark
is already lifted, and compute  the mean of the
conditional Gaussian distribution  $p(l_i|l_{\mathbf{pa}(l_i)})$ of $l_i$
conditioned on its limb parent $\mathbf{pa}(l_i)$ to further compute $\bar{x_i}$.

Note that our approach requires some body landmarks to be detected.
Indeed,  as in our opinion it is unrealistically to attempt determining
the complete 3D pose of the person from a few detected body landmarks, \eg the arm, 
%
we assume that at least the spine limb and other two
body landmarks in the trunk (shoulders, heaps) are detected.

\section{Human 3D Pose Estimation}
\label{sec:3dpose}
%
%
This section presents our residual-pose learning approach to predict (in a camera coordinate frame )
the 3D coordinates of a human  skeleton comprising $\NJoints$ body landmarks.
%

\subsection{Residual Pose Learning}
Provided the 2D body landmark detections, our lifting step provides
a \emph{rough} estimate of the  3D pose.
%
Yet, lifted values will exhibit 3D pose estimation errors,
specially since lifted 3D points lie on the depth surface rather
than represent the inner joint (see Fig.~\ref{fig:calibrated-skeleton}).
In this regard, in absence of other sources of errors (missed detections,
occlusion, etc.) we can argue that such estimates differ only
from the true 3D pose by some coordinate offset.
This inspired us to follow a simple yet effective approach to obtain
refined estimates from rough lifted estimates.

\begin{figure}[t]
\centering
\includegraphics[width=\linewidth]{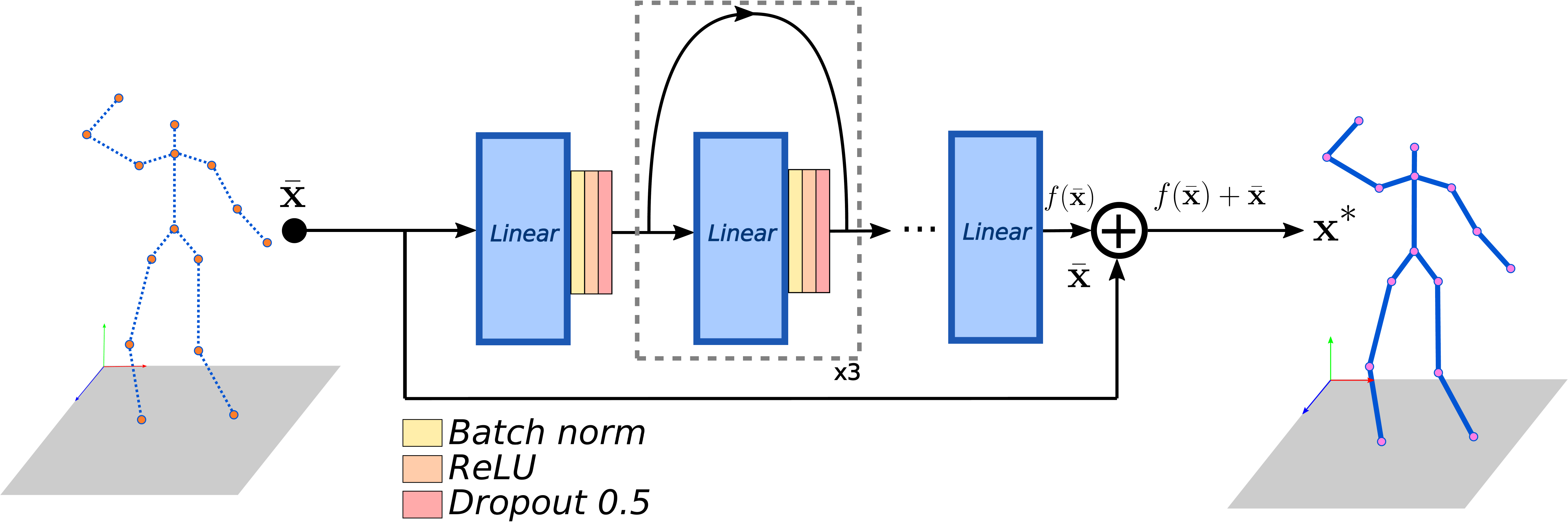}
\vspace{-0.2cm}
\caption{
  Residual pose learning framework.
  Our neural network regressor receives as input a lifted 3D pose $\liftedpose$.
  Due to the global skip connection, the regressor has to
  predict the residual pose $f(\liftedpose)$ to be added to $\liftedpose$
  to predict the true 3D pose. 
  %
  %
  The building block of our neural network regressor is a linear layer
  followed by batch normalization, ReLU activations and dropout, and 
  with a skip connection.
}
\vspace{-2mm}
\label{fig:residual_pose}
\end{figure}

Our approach can be set as follows: given a \emph{rough} 3D pose 
estimate $\liftedpose\in \mathbb{R}^{\NJoints\times 3}$
obtained from the 2D landmark detection lifting step, 
%
and its \emph{true} corresponding 3D pose $\truepose\in\mathbb{R}^{\NJoints\times 3}$, 
the neural regressor $f$ can focus on modelling their residual $\truepose-\liftedpose$ as:
\begin{equation}
f(\liftedpose)+\liftedpose=\truepose.
\label{eq:residualpose}
\end{equation}
The function $f(\liftedpose)$ is the residual to be learned.
Graphically, these residuals represent the vector of coordinate offsets that are necessary
to predict the true 3D pose $\truepose$ (hence a residual pose).
Architecturally speaking, the operation $f(\liftedpose)+\liftedpose$
is performed by a shortcut connection with the identity mapping of $\liftedpose$,
as shown in  Fig.~\ref{fig:residual_pose}. 
%

%
%


Additionally, we can augment $\liftedpose$ by incorporating the confidence
of the 2D detections provided by the 2D pose estimation CNN.
This will add an extra dimension for each detected landmark 
$\liftedpose\in\mathbb{R}^{J\times 4}$.
In such case the shortcut connection works as a pooling layer that removes  the
extra dimension to match the one of $\truepose$.
We analyze this particular case in Section~\ref{sec:experiments}.




%

\subsection{Neural Network Regressor}
\label{sec:regarch}
%
%
We aim to find a simple and efficient network architecture~$f$ that
performs well enough in the regression task.
Fig.\ref{fig:residual_pose} shows a diagram with the basic building
blocks of our architecture.
It is a multi-layer network consisting on a series of fully-connected
layers, each followed by batch normalization, ReLU activations and
dropout layers.
The first layer receives as input the lifted pose $\liftedpose$
and outputs $\NFeatures$ features.
This number of features are kept fixed until the output layer that 
generates the residual pose vector in $\mathbb{R}^{\NJoints\times 3}$.
Each of the inner layers have skip connections.
One can normally squeeze as many inner layers $\RegStages$ to make the 
regressor deeper.
However, in this paper we set $\RegStages=3$.

\subsection{Pose Learning Loss}
Let $\totalpose=f(\liftedpose)+\liftedpose$ be the 3D pose prediction.
We use the following loss to train our neural network regressor
\begin{equation}
L_{res} = \frac{1}{\NJoints}\sum^{\NJoints}_{i=1} || \totalpose_i - \truepose_i ||_1,
\label{eq:jointloss}
\end{equation}
where $\truepose_i$ is the ground truth of the body landmark $i$ and 
$ \totalpose_i$ is the 3D prediction for such landmark.
In our experiments we use the smooth L1 norm as we found out that it
works better than the L2 or plain L1 norms.
%

%

\section{Experiments}
\label{sec:experiments}
We conducted several experiments to evaluate our approach  effectiveness
in single and multi-person scenarios.

\subsection{Depth-image datasets}

\mypartitle{ITOP~\cite{DepthInvariant}}.
%
%
This  dataset consists of images in a single person pose estimation setting.
It has 18k and 5k depth images for training and testing, respectively, recorded
with an Asus Xtion camera.
It  was built from 20 subjects performing 15 different actions each.

\mypartitle{CMU-Panoptic~\cite{Joo_2017_TPAMI}}.
It comprises multiple recordings acquired with different sensor devices such as
color and depth cameras (Kinect2). 
We consider a subset of the depth recordings from the \emph{Haggling} category.
The setup contains several interacting people with diverse body pose configurations
with respect to the camera and between-person interactions.
For training we selected 15k 3D person instances from the sequence 
170407\underline{ }haggling\underline{ }a3 for training.
For testing 1.5k 3D person instances were selected 
from the sequence 170407\underline{ }haggling\underline{ }b3.

\subsection{Evaluation metrics}
\mypartitle{Mean average precision (mAP)}.
As standard practice in 3D human pose estimation, we use mean average precision at 10 cm (mAP{@}10cm) to measure the 
3D detection performance. 
A successful detection is considered when the detected 3D body landmark
falls within a distance less than 10~cm from the ground truth.  
We report the average precision~(AP) for individual body landmarks
and to measure the overall performance,
the mean average precision~(mAP) defined as the mean of the APs of all body landmarks. 
Larger values are better.

\mypartitle{Mean per joint position error~(MPJPE)}.
It measures the average error in Euclidean distance between the 
detected 3D body landmarks and the ground truth.
Lower values are better.
We report MPJPE in  centimeters for each body landmark
and their mean for the overall performance~(mMPJPE).

\mypartitle{Percentage of correct keypoints~(PCK)}.
We use PCK to evaluate the performance of the 2D pose estimation task.
It relies in the precision and recall that result from the percentage
of correct detected keypoints (body landmarks).
We follow the evaluation protocol presented in~\cite{Martinez_IROS_2018}.
For each joint (\eg knee), true positives, false positives, and false negatives
are counted using a radius obtained according to the height of the bounding box
(ground truth) containing the person.
Then, the precision and recall rates are calculated by averaging the above
values over a set of varying radius, body landmarks, and dataset samples.

\subsection{Implementation details}

\mypartitle{Image pre-processing}.
We normalize the depth images by linearly scaling the depth sensor values in
$[0,8]$~meter range into the  $[-0.5,0.5]$ range.

\mypartitle{2D CNN architectures and training}.
We keep the performance-efficiency trade-off reported in \cite{Martinez_TCSVT_2019} and
experiment with RPM with 2 stages and MPM with 4 stages.
We configure the Hourglass architecture (HG) to 2 stages as it was shown
that performance saturates at this point~\cite{HourGlass}.

We train the 2D pose estimation CNNs using Adam.
To avoid overfiting due to the low number of depth images in the addressed
datasets, and increase the 2D pose performance, we train the networks for 
13 epochs with the large synthetic people dataset introduced 
in~\cite{Martinez_IROS_2018}. 
%
%
Then, the CNNs are finetuned using the real dataset (ITOP or CMU-Panoptic)
for 100 epochs.

\mypartitle{Residual Pose Regressor}.
We train our neural network regressor for 200 epochs using Adam and 
minibatches of size 128.
We apply standard normalization to the 3D lifted pose and the 3D ground truth 
pose by substracting the mean and dividing by the standard deviation.
We select $1e-3$ as initial learning rate and decrease it by 2 every 20 epochs.

\subsection{Experimental results}

\begin{figure}
\includegraphics[width=0.49\linewidth]{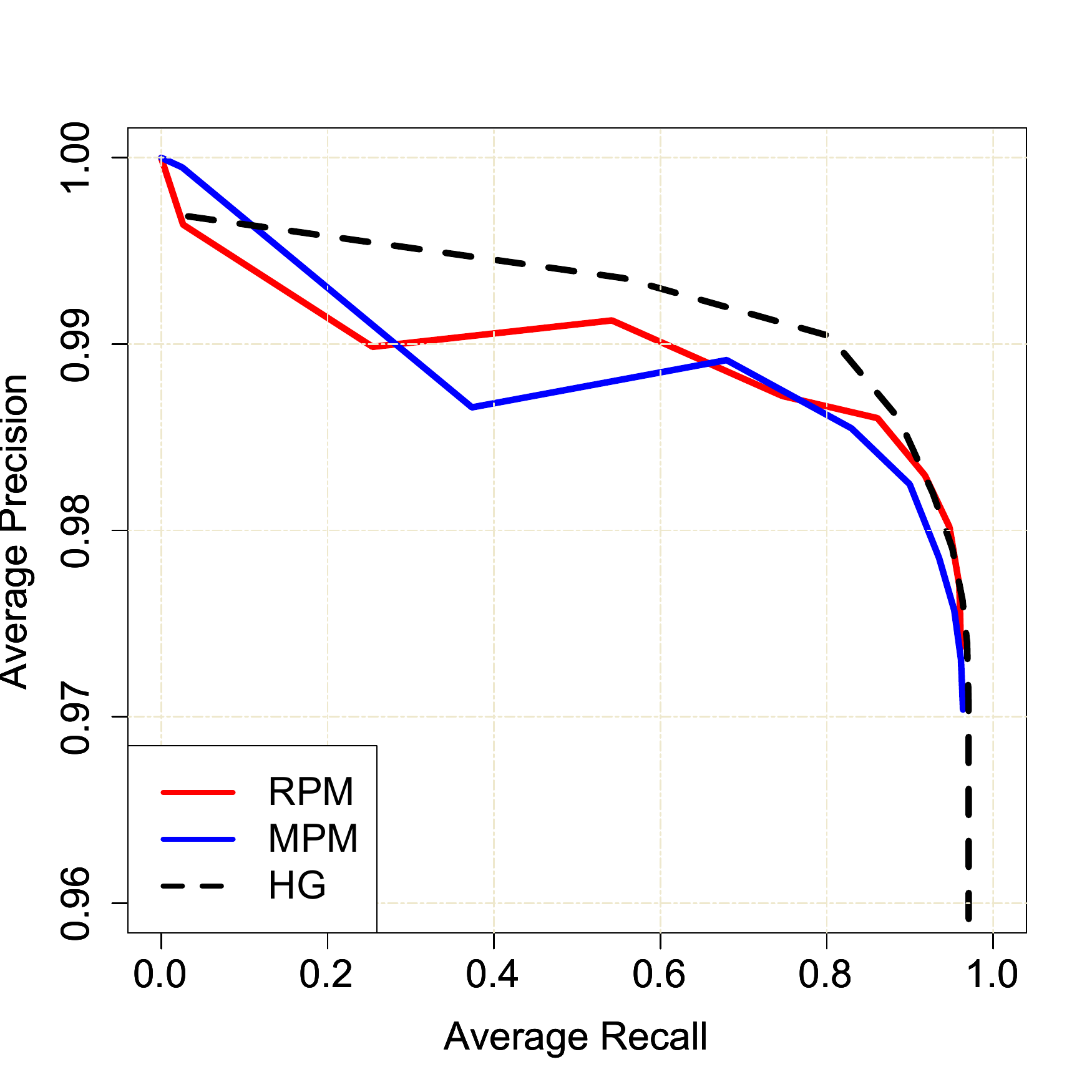}
\includegraphics[width=0.49\linewidth]{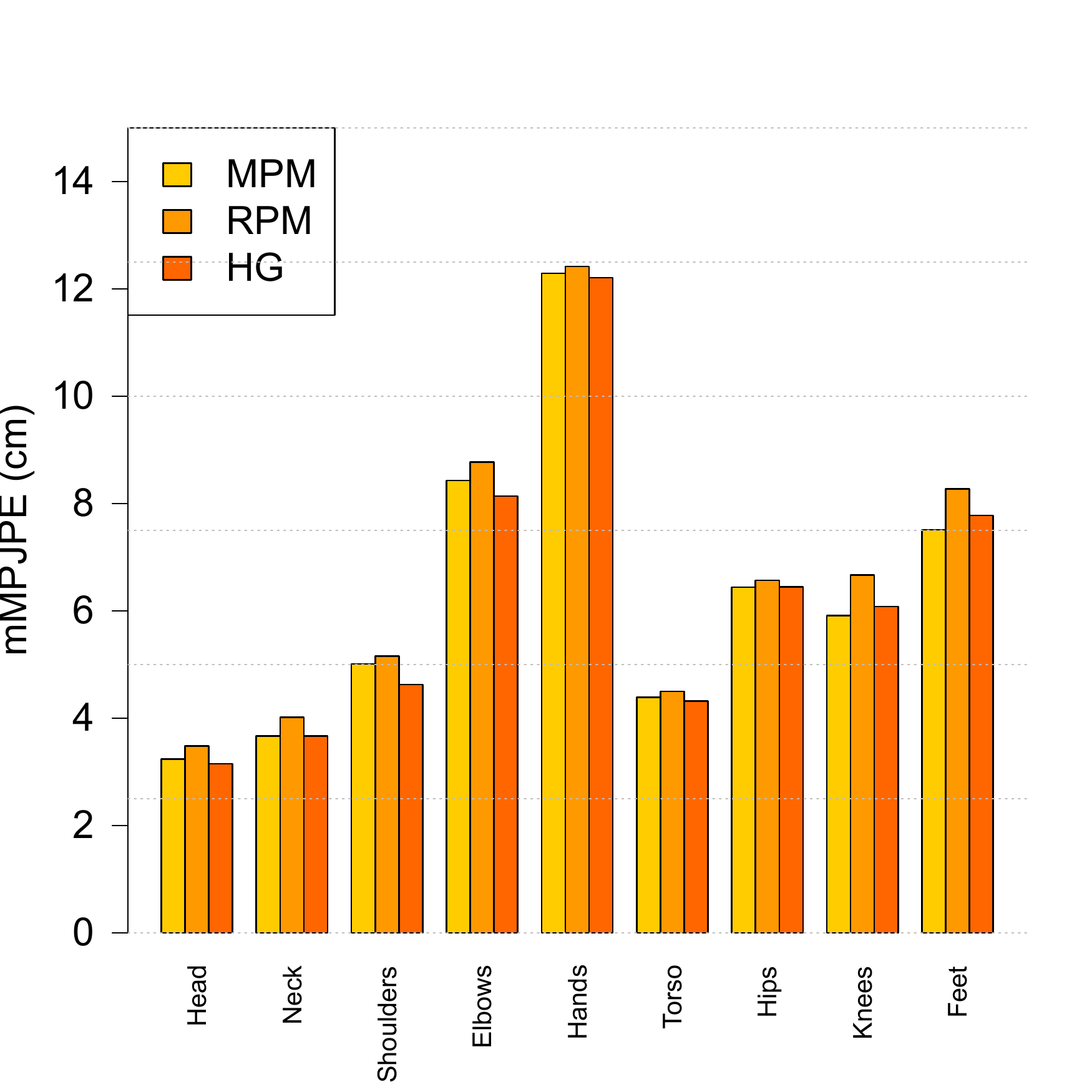}

\vspace{-0.2cm}
\caption{Performance of the different CNNs for 2D pose estimation. 
  Left: 2D pose estimation performance measured with recall and precision curves.
  Right: resulting 3D estimation pose performance in terms of MPJPE for each 
  body part. The lower the better.
}
\vspace{-2mm}
\label{fig:curves}
\end{figure}

\begin{figure*}[t]
\centering
\includegraphics[width=\textwidth]{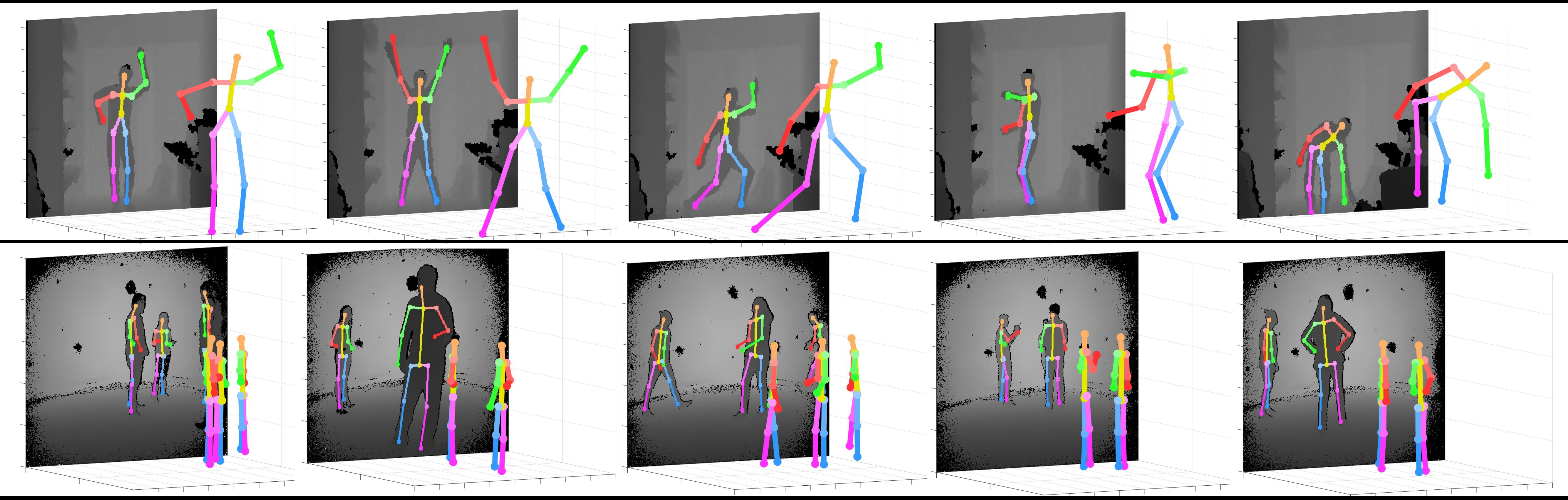}
	\vspace{-0.2cm} 
	\caption{3D pose estimation examples and their 2D projection
	of our approach on the single person ITOP dataset 
	(top row) and the multi-person CMU-Panoptic dataset (bottom row). 	
	}

\label{fig:3d-skeletons}
\end{figure*}

\begin{table}
\centering

\begin{tabular}{l | c c c }
\Xhline{2\arrayrulewidth}

CNN model   & MPM    & RPM    & HG   \\

\Xhline{2\arrayrulewidth}

FPS         & 84    &  35     & 18  \\ 

\# Params   & 304.9K &  2.84M & 12.9M\\

\Xhline{2\arrayrulewidth}

F-Score (2D)& 0.96   &  0.96  & \textbf{0.97} \\ \hline

mAP{@}10cm & 85.61  &  85.96  & \textbf{85.97}\\

mMPJPE      & 6.83   & 7.18    & \textbf{6.78}  \\
\Xhline{2\arrayrulewidth}

\end{tabular}

\caption{2D and 3D pose estimation performance obtained
		for the different 2D CNN architectures and their
		computational complexity.
}
\label{tab:efficiency}
\end{table}

\begin{table*}[t]
\centering

\begin{tabular}{l | c c c c c c c c }

\Xhline{2\arrayrulewidth}

         & \multicolumn{8}{c}{ITOP (front-view)} \\

\Xhline{2\arrayrulewidth}

	& \multicolumn{8}{c}{AP{@}10cm} \\

\Xhline{2\arrayrulewidth}

Body part & \cite{Jung_RTW_CVPR_2015} & \cite{DepthInvariant} & \cite{guo2017towards} & \cite{Moon_2018_CVPR_V2V-PoseNet} & \textbf{\ResPose} &\textbf{\ResPose}$^{*}$& \textbf{\ResPose}$^{-}$& \CoordReg     \\

\Xhline{2\arrayrulewidth}
\rowcolor{Gray}
Head      & 97.8 & 98.1 & \textbf{98.7} & \underline{98.29} & 98.27  &  98.13  & 98.33 & 97.8  \\
Neck      & 95.8 & 97.5 & \textbf{99.4} & \underline{99.07} & 98.6   &  98.56  & 98.5 & 98.66  \\
\rowcolor{Gray}
Shoulders & 94.1 & \underline{96.5} & 96.1 & \textbf{97.18} & 95.34  &  95.2   & 92.78 & 95.64 \\
Elbows    & \underline{77.9} & 73.3 & 74.7 & \textbf{80.42} & 76.52  &  75.89  & 74.38 & 74.24 \\
\rowcolor{Gray}
Hands     & \textbf{70.5} & \underline{68.7} & 55.2 & 67.26 & 61.69  &  61.28  & 59.98 & 55.01 \\
Torso     & 93.8 & 85.6 & \underline{98.7} & \textbf{98.73} & 98.56  &  98.64  & 98.62 & 97.57 \\
\rowcolor{Gray}
Hips      & 80.3 & 72   & \underline{91.8} & \textbf{93.23} & 90.07  &  90.31  & 89.4 & 87.09  \\
Knees     & 68.8 & 69   & 89   & \textbf{91.80} & \underline{89.13}  &  88.93  & 88.82 & 88.29 \\
\rowcolor{Gray}
Feet      & 68.4 & 60.8 & 81.1 & \textbf{87.6}  & \underline{84.28}  &  83.52  & 83.66 & 83.99 \\

\Xhline{2\arrayrulewidth}

Mean (mAP)     & 80.5 & 77.4 & 84.9 & \textbf{88.74} & \underline{85.97}  &  85.71  & 84.9 & 84.17  \\

\Xhline{2\arrayrulewidth}

          & \multicolumn{8}{c}{CMU-Panoptic} \\

\Xhline{2\arrayrulewidth}

	     & \multicolumn{4}{c|}{MPJPE (cm)} &  \multicolumn{4}{c}{AP{@}10cm}  \\
	     
\Xhline{2\arrayrulewidth}
	     
Body part & \textbf{\ResPose} & \textbf{\ResPose}$^{*}$ & \textbf{\ResPose}$^{-}$ & \multicolumn{1}{c|}{\CoordReg} &  \textbf{\ResPose} & \textbf{\ResPose}$^{*}$ & \textbf{\ResPose}$^{-}$ & \CoordReg \\

\Xhline{2\arrayrulewidth}
\rowcolor{Gray}
Head      & \textbf{6.59}  & \underline{6.78}  &  10.17  & \multicolumn{1}{c|}{11.17} & \underline{96.4} &  \textbf{96.67}  & 79.47  & 72.33 \\
Neck      & \textbf{7.29}  & \underline{7.45}  &  8.5    & \multicolumn{1}{c|}{11.68} & \textbf{96.53}  & \underline{96.2} & 92.13  & 74.07 \\
\rowcolor{Gray}
Shoulders & \textbf{8.55}  &  \underline{8.66} &  10.96  & \multicolumn{1}{c|}{14.38} & \textbf{87.17}  & \underline{85.6}   & 77.17  &54.33  \\
Elbows    & \underline{14.52} &  \textbf{14.19} &  23.86  & \multicolumn{1}{c|}{20.2}  & \underline{59.17}  &  \textbf{61.97}  & 38.3   & 28.93 \\
\rowcolor{Gray}
Hands     & \textbf{27.85} &  \underline{27.96} &  31.16  & \multicolumn{1}{c|}{26.37} & 16.63  & \underline{17.47}& \textbf{17.77}  & 6.37  \\
Torso     & \underline{9.06}  &  \textbf{8.51}  &  9.92   & \multicolumn{1}{c|}{11.93} & \textbf{93.27}  & \underline{92.67} & 87.6   &67.53  \\
\rowcolor{Gray}
Hips      & \textbf{8.57}  &  \underline{8.67}  &  12.16  & \multicolumn{1}{c|}{12.99} & \textbf{91.97}  & \underline{90.27} & 70.1   &66.1   \\
Knees     & \textbf{9.24}  &  \underline{9.43}  &  14.72  & \multicolumn{1}{c|}{13.96} & \textbf{81.8}   & \underline{80.6} & 58.67  &52.33  \\
\rowcolor{Gray}
Feet      & \underline{11.26} &  \textbf{11.19} &  18.8   & \multicolumn{1}{c|}{15.54} & \textbf{70.77}  & \underline{70.5}   & 52.17  & 48.27 \\

\Xhline{2\arrayrulewidth}

Mean      & \textbf{12.2}  &  \textbf{12.2}  &  16.79  & \multicolumn{1}{c|}{16.11} & \textbf{73.41}  & \underline{73.22} & 59.17  & 48.44 \\

\Xhline{2\arrayrulewidth}

\end{tabular}


\caption{3D pose estimation performance. 
		 Top: mAP of the state-of-the-art on single person pose estimation setting in the 
		 ITOP dataset,
		 Bottom: mAP and mMPJPE for the multi-person pose estimation setting in the 
		 CMU-Panoptic dataset.
}
\label{tab:soa-map}
\end{table*}

\mypartitle{2D Pose Network Architectures}.
We evaluate the quality of the 2D pose predictions for the 3D pose estimation
task in the ITOP dataset.
Fig.~\ref{fig:curves} shows the 2D pose estimation performance curves and the
3D pose error in terms of MPJPE for the different CNNs.
Table~\ref{tab:efficiency} summarizes these results with the maximum F-Score
obtained for 2D pose estimation, and the mAP and mMPJPE for 
3D pose prediction.
Indeed, providing better 2D pose estimates reflects directly in the 3D
performance.
Overall the HG 2D detections provide the best 3D estimates
achieving the lowest mMPJPE and better mAP.
%
We select the HG network for the rest of the analysis.

\mypartitle{Computational Requirements}.
Table~\ref{tab:efficiency} reports the number of parameters of each CNN
and the frames per second (FPS) required for the forward pass
in a single Nvidia card GTX 1050.
Note that the FPS is also valid for the multi-person case since the CNNs
predict the pose for each individual in the image in a single forward pass.
Additionally, the neural network regressor requires 12.7M parameters but runs 
at 1700 FPS, so its cost, even when applied for multiple person,  is negligible 
compared to that of a 2D pose CNN.
Hence our proposed approach can run very 
efficiently in real-time in a single GPU.

\mypartitle{Comparison with the state of the art}.
%
%
%
Table~\ref{tab:soa-map} compares the detailed AP scores for each body landmark
of our proposed approach (\textbf{\ResPose}) with the state of the art in 
the ITOP dataset.
Overall our residual pose learning approach shows very competitive results 
obtaining the second best performance.
%
%
The best performing work is~\cite{Moon_2018_CVPR_V2V-PoseNet} that processes
voxelized representations of the 3D space processed with a 3D CNN,
and uses an ensemble of 10 models for the final prediction.
Contrary, our residual pose approach is simpler and efficient. 
%
%
Example results are shown in Fig.~\ref{fig:3d-skeletons} (top row).

\mypartitle{Multi-person 3D pose estimation}.
Table~\ref{tab:soa-map} reports AP and MPJPE for the
multi-person setting in the CMU-Panoptic dataset.
Naturally the ranges of pose profile, multiple scales, and the quality of sensing
make this setup more challenging than the single person pose setting.
The more affected body landmarks are the hands and elbows with lower AP and 
larger MPJPE.
Note these are the elements that are in constant motion and are more affected 
by self occlusions,
compared to other elements like the torso and head.
Fig.~\ref{fig:3d-skeletons} shows prediction examples.

\mypartitle{Recovery from 2D Failures}.
We report the results of removing the prior recovery component introduced 
in Section~\ref{sec:lifting}.
Table~\ref{tab:soa-map} shows the performance for the single and multi-person
settings (\textbf{\ResPose}$^{-}$).
The performance drops specially for the multi-person scenario.
%

\mypartitle{2D Landmark Detection Confidence}.
We incorporated the confidence of the 2D detections provided by the CNN
that range in $[0,1]$ in our residual pose learning setting.
When a landmark was recovered by the process of Section~\ref{sec:lifting},
we set a low confidence value of $\sigma=0.1$ to identify them from the rest.
The results are reported in Table~\ref{tab:soa-map} (\textbf{\ResPose}$^{*}$).
The mAP slightly decreases in this case.
However, in the multi-person setting some especific elements 
(head, elbows, hands) have slightly better detection rate.

\mypartitle{Coordinate Regression}.
We experimented with 3D coordinate regression using the neural
network architecture introduced in Section~\ref{sec:regarch} and 
predict $X,Y,Z$  coordinates of the body landmarks from 
lifted 2D detections, dropping the residual pose connection.
Table~\ref{tab:soa-map} compare these results (\CoordReg) with our residual
pose approach.
The performance drops for both single and multi-person settings.
Certainly when people appear roughly in the same position, as
the case in ITOP dataset, 3D coordinate regression presents
a good alternative.
However, our residual pose approach outperforms 3D coordinate
regression in both, single and multi-person settings.

\section{Conclusions} \label{sec:conclusion}
%
%
%

In this paper we addressed the problem of 3D pose estimation from depth images.
We decoupled 2D and 3D pose estimation and predict the 3D pose from lifted
2D detections.
We proposed a residual-pose regression learning to predict the 3D pose
by refining lifted detections.
We introduced a pairwise 3D limb prior to recover from 2D detection failures
and analyse the incorporation of 2D detection confidence in our pipeline.
Despite the simplicity of our approach we achieve competitive results in two 
public datasets for single and multi-person pose estimation.
Our method propose
a more efficient alternative for multi-party HRI settings.

Our study opens the way for new research.
One limitation of our model is that it does not consider the skeleton
kinematics in the learning process.
Additionally, body motion modelling can be introduced to introduce
temporal consistency in our 3D predictions.

\mypartitle{Acknowledgments}:
This work was supported by the European Union under the EU Horizon 2020 Research
and Innovation Action MuMMER (MultiModal Mall Entertainment Robot), project ID
688146, as well as the Mexican National Council for Science and Technology~(CONACYT)
under the PhD scholarships~program.

\bibliographystyle{plain}
\bibliography{biblio}

\begin{thebibliography}{10}

\bibitem{Aksan_2019_ICCV}
Emre Aksan, Manuel Kaufmann, and Otmar Hilliges.
\newblock Structured prediction helps 3d human motion modelling.
\newblock In {\em The IEEE International Conference on Computer Vision (ICCV)},
  Oct 2019.

\bibitem{KEEPSMPL_ECCV_2016}
Federica Bogo, Angjoo Kanazawa, Christoph Lassner, Peter Gehler, Javier Romero,
  and Michael~J. Black.
\newblock Keep it smpl: Automatic estimation of 3d human pose and shape from a
  single image.
\newblock In {\em Computer Vision -- ECCV 2016}, pages 561--578, 2016.

\bibitem{CPMPaf}
Zhe Cao, Tomas Simon, Shih-En Wei, and Yaser Sheikh.
\newblock Realtime multi-person 2d pose estimation using part affinity fields.
\newblock In {\em CVPR}, 2017.

\bibitem{Chen_2017_CVPR}
C.~{Chen} and D.~{Ramanan}.
\newblock 3d human pose estimation = 2d pose estimation + matching.
\newblock In {\em 2017 IEEE Conference on Computer Vision and Pattern
  Recognition (CVPR)}, pages 5759--5767, July 2017.

\bibitem{guo2017towards}
Hengkai Guo, Guijin Wang, Xinghao Chen, and Cairong Zhang.
\newblock Towards good practices for deep 3d hand pose estimation.
\newblock {\em arXiv preprint arXiv:1707.07248}, 2017.

\bibitem{Habibie_CVPR_2019}
Ikhsanul Habibie, Weipeng Xu, Dushyant Mehta, Gerard Pons-Moll, and Christian
  Theobalt.
\newblock In the wild human pose estimation using explicit 2d features and
  intermediate 3d representations.
\newblock In {\em IEEE Conf. on Computer Vision and Pattern Recognition
  (CVPR)}, 2019.

\bibitem{DepthInvariant}
Albert Haque, Boya Peng, Zelun Luo, Alexandre Alahi, Serena Yeung, and
  Li~Fei-Fei.
\newblock Towards viewpoint invariant 3d human pose estimation.
\newblock In {\em European Conference on Computer Vision (ECCV)}, 2016.

\bibitem{Joo_2017_TPAMI}
Hanbyul Joo, Tomas Simon, Xulong Li, Hao Liu, Lei Tan, Lin Gui, Sean Banerjee,
  Timothy~Scott Godisart, Bart Nabbe, Iain Matthews, Takeo Kanade, Shohei
  Nobuhara, and Yaser Sheikh.
\newblock Panoptic studio: A massively multiview system for social interaction
  capture.
\newblock {\em IEEE Transactions on Pattern Analysis and Machine Intelligence},
  2017.

\bibitem{Jung_RTW_CVPR_2015}
Ho~Yub Jung, Soochahn Lee, Yong~Seok Heo, and Il~Dong Yun.
\newblock Random tree walk toward instantaneous 3d human pose estimation.
\newblock In {\em CVPR}, pages 2467--2474. IEEE Computer Society, 2015.

\bibitem{kanazawaHMR_CVPR_18}
Angjoo Kanazawa, Michael~J. Black, David~W. Jacobs, and Jitendra Malik.
\newblock End-to-end recovery of human shape and pose.
\newblock In {\em Computer Vision and Pattern Regognition (CVPR)}, 2018.

\bibitem{Li_ICCV_2015}
Sijin Li, Weichen Zhang, and Antoni~B. Chan.
\newblock Maximum-margin structured learning with deep networks for 3d human
  pose estimation.
\newblock In {\em IEEE International Conference on Computer Vision (ICCV)},
  2015.

\bibitem{martinez_2017_3dbaseline}
Julieta Martinez, Rayat Hossain, Javier Romero, and James~J. Little.
\newblock A simple yet effective baseline for 3d human pose estimation.
\newblock In {\em ICCV}, 2017.

\bibitem{Martinez_IROS_2018}
Angel Mart\'inez-Gonz\'alez, Michael Villamizar, Olivier Can{\'e}vet, and
  Jean-Marc Odobez.
\newblock Real-time convolutional networks for depth-based human pose
  estimation.
\newblock In {\em 2018 {IEEE/RSJ} International Conference on Intelligent
  Robots and Systems, {IROS}}, 2018.

\bibitem{Martinez_TCSVT_2019}
Angel {Martínez-González}, Michael {Villamizar}, Olivier {Canévet}, and
  Jean-Marc {Odobez}.
\newblock Efficient convolutional neural networks for depth-based multi-person
  pose estimation.
\newblock {\em IEEE Transactions on Circuits and Systems for Video Technology},
  2019.

\bibitem{Moon_2018_CVPR_V2V-PoseNet}
Gyeongsik Moon, Juyong Chang, and Kyoung~Mu Lee.
\newblock V2v-posenet: Voxel-to-voxel prediction network for accurate 3d hand
  and human pose estimation from a single depth map.
\newblock In {\em The IEEE Conference on Computer Vision and Pattern
  Recognition (CVPR)}, 2018.

\bibitem{HourGlass}
Alejandro Newell, Kaiyu Yang, and Jia Deng.
\newblock Stacked hourglass networks for human pose estimation.
\newblock In {\em European Conference on Computer Vision}, 2016.

\bibitem{pavlakos2017volumetric}
Georgios Pavlakos, Xiaowei Zhou, Konstantinos~G Derpanis, and Kostas
  Daniilidis.
\newblock Coarse-to-fine volumetric prediction for single-image 3{D} human
  pose.
\newblock In {\em Proceedings of the IEEE Conference on Computer Vision and
  Pattern Recognition}, 2017.

\bibitem{Ramakrishna_ECCV_2012}
Varun Ramakrishna, Takeo Kanade, and Yaser Sheikh.
\newblock Reconstructing 3d human pose from 2d image landmarks.
\newblock In {\em European Conference on Computer Vision}, 2012.

\bibitem{ShottonPAMI}
Jamie Shotton, Ross Girshick, Andrew Fitzgibbon, Toby Sharp, Mat Cook, Mark
  Finocchio, Richard Moore, Pushmeet Kohli, Antonio Criminisi, Alex Kipman, and
  Andrew Blake.
\newblock Efficient human pose estimation from single depth images.
\newblock {\em Trans. PAMI}, January 2012.

\bibitem{Sigal:IJCV:11}
L.~Sigal, M.~Isard, H.~Haussecker, and M.~J. Black.
\newblock Loose-limbed people: Estimating {3D} human pose and motion using
  non-parametric belief propagation.
\newblock {\em International Journal of Computer Vision}, 98(1):15--48, May
  2011.

\bibitem{Sun2017CompositionalHP}
Xiao~Wei Sun, Jiaxiang Shang, Shuang Liang, and Yichen Wei.
\newblock Compositional human pose regression.
\newblock In {\em IEEE International Conference on Computer Vision (ICCV)},
  2017.

\bibitem{Taylor_Viturvian_CVPR_2018}
Jonathan Taylor, Jamie Shotton, Toby Sharp, and Andrew~W. Fitzgibbon.
\newblock The vitruvian manifold: Inferring dense correspondences for one-shot
  human pose estimation.
\newblock In {\em CVPR}, 2012.

\bibitem{Tome_CVPR_2017}
D.~{Tome}, C.~{Russell}, and L.~{Agapito}.
\newblock Lifting from the deep: Convolutional 3d pose estimation from a single
  image.
\newblock In {\em IEEE Conf. on Computer Vision and Pattern Recognition
  (CVPR)}, 2017.

\bibitem{Wandt_CVPR_2019}
Bastian Wandt and Bodo Rosenhahn.
\newblock Repnet: Weakly supervised training of an adversarial reprojection
  network for 3d human pose estimation.
\newblock In {\em CVPR}, 2019.

\bibitem{DepthMultitask}
Keze Wang, Shengfu Zhai, Hui Cheng, Xiaodan Liang, and Liang Lin.
\newblock Human pose estimation from depth images via inference embedded
  multi-task learning.
\newblock In {\em Proceedings of ACM on Multimedia}, 2016.

\bibitem{ICP_ICCV_2011}
M.~{Ye}, {Xianwang Wang}, R.~{Yang}, {Liu Ren}, and M.~{Pollefeys}.
\newblock Accurate 3d pose estimation from a single depth image.
\newblock In {\em 2011 International Conference on Computer Vision}, pages
  731--738, Nov 2011.

\end{thebibliography}

\end{document}